\documentclass[preprint,12pt]{elsarticle}
\usepackage{graphicx}%
\usepackage{multirow}%
\usepackage{amsmath,amssymb,amsfonts}%
\usepackage{amsthm}%
\usepackage{mathrsfs}%
\usepackage[title]{appendix}%
\usepackage{xcolor}%
\usepackage{textcomp}%
\usepackage{manyfoot}%
\usepackage{booktabs}%
\usepackage{algorithm}%
\usepackage{algorithmicx}%
\usepackage{algpseudocode}%
\usepackage{listings}%
\usepackage{colortbl}%
\usepackage{xcolor}

\usepackage{booktabs}
\usepackage{algorithm}
\usepackage{algpseudocode}
\usepackage{hyperref}
\usepackage{amsmath}
\usepackage{array}%
\usepackage{stfloats}%
\usepackage{url}%
\usepackage{tabularx}%
\usepackage{verbatim}%
\usepackage{longtable}%
\usepackage{lscape}%
\usepackage{pifont}%
\usepackage{adjustbox}%
\usepackage{tikz}%
\usepackage[caption=false,font=normalsize,labelfont=sf,textfont=sf]{subfig}%
\usepackage{enumitem}%
\usepackage{microtype}%
\usepackage{mathtools,xspace}%
\usepackage{bm}%
\usepackage{natbib} %

\newcommand{\onedot}{\ifx\@let@token.\else.\null\fi\xspace}
\newcommand{\etal}{\emph{et al}\onedot}

\newcommand{\ie}{\emph{i.e}\onedot}

\theoremstyle{thmstyleone}

\theoremstyle{thmstyletwo}%

\theoremstyle{thmstylethree}%

\usepackage{amssymb}

\usepackage{amsmath}
\usepackage{xcolor}
\usepackage[normalem]{ulem}

\journal{Nuclear Physics B}

\begin{document}

\begin{frontmatter}

\title{Adapting Segment Anything Model for Power Transmission Corridor Hazard Segmentation}

\author{%
  Hang Chen\textsuperscript{1, \textdagger},%
  \quad Maoyuan Ye\textsuperscript{1, \textdagger},%
  \quad Peng Yang\textsuperscript{1},%
  \quad Haibin He\textsuperscript{1},%
  \\ Juhua Liu\textsuperscript{1, \textsection},%
  \quad Shaohe Wang\textsuperscript{2},
  \quad Bo Du\textsuperscript{1}\\[4pt]
  {\small \{chenhang,yemaoyuan,pengyang,haibinhe,liujuhua,dubo\}@whu.edu.cn}\\
  volk-hyllow@hotmail.com \\
  [2pt]
  {\small \textsuperscript{\textdagger} These authors contributed equally to this work}\\[-2pt]
  {\small \textsuperscript{\textsection} Corresponding author}%
}

\address[1]{School of Computer Science, National Engineering Research Center for Multimedia Software, Institute of Artificial Intelligence, and Hubei Key Laboratory of Multimedia and Network Communication Engineering, Wuhan University, Wuhan, China}
\address[2]{Institute of Power Transmission and Transformation Technology, State Grid Zhejiang Electric Power Co., LTD, Research Institute Hangzhou, China.}

\begin{abstract}
Power transmission corridor hazard segmentation (PTCHS) aims to separate transmission equipment and surrounding hazards from complex background, conveying great significance to maintaining electric power transmission safety.
Recently, the Segment Anything Model (SAM) has emerged as a foundational vision model and pushed the boundaries of segmentation tasks. 
However, SAM struggles to deal with the target objects in complex transmission corridor scenario, especially those with fine structure.
In this paper, we propose ELE-SAM, adapting SAM for the PTCHS task. Technically, we develop a Context-Aware Prompt Adapter to achieve better prompt tokens via incorporating global-local features and focusing more on key regions. Subsequently, to tackle the hazard objects with fine structure in complex background, we design a High-Fidelity Mask Decoder by leveraging multi-granularity mask features and then scaling them to a higher resolution. Moreover, to train ELE-SAM and advance this field, we construct the ELE-40K benchmark, the first large-scale and real-world dataset for PTCHS including 44,094 image-mask pairs. Experimental results for ELE-40K demonstrate the superior performance that ELE-SAM outperforms the baseline model with the average 16.8\% mIoU and 20.6\% mBIoU performance improvement. Moreover, compared with the state-of-the-art method on HQSeg-44K, the average 2.9\% mIoU and 3.8\% mBIoU absolute improvements further validate the effectiveness of our method on high-quality generic object segmentation. 
The source code and dataset are available at \url{https://github.com/Hhaizee/ELE-SAM}.
\end{abstract}


\end{frontmatter}

\section{Introduction}
Power transmission corridor inspection is crucial for maintaining the electric power transmission safety and stability~\citep{9802679,9699431,8972612}. Current inspection methods based on object detection have achieved great progress in locating hazard targets around transmission corridors~\citep{LI2024123983,9530535}. However, they typically predict rough bounding boxes, hindering subsequent tasks (morphology analysis, ranging) that require accurate hazard shape.
Therefore, power transmission corridor hazard segmentation (PTCHS) aims to further separate transmission equipment and surrounding hazards from complex background, providing a more practical solution for the following risk assessment and prevention strategy development. While PTCHS does not introduce a fundamentally new segmentation paradigm, it addresses the critical real-world need for hazard screening in power transmission corridors. Related studies have achieved promising results~\citep{9699431,Zhou2022DualViewSA}. Nevertheless, the lack of precise segmentation limits the practical effectiveness of PTCHS.

\begin{figure}[!t]
    \centering
\includegraphics[width=0.8\linewidth]{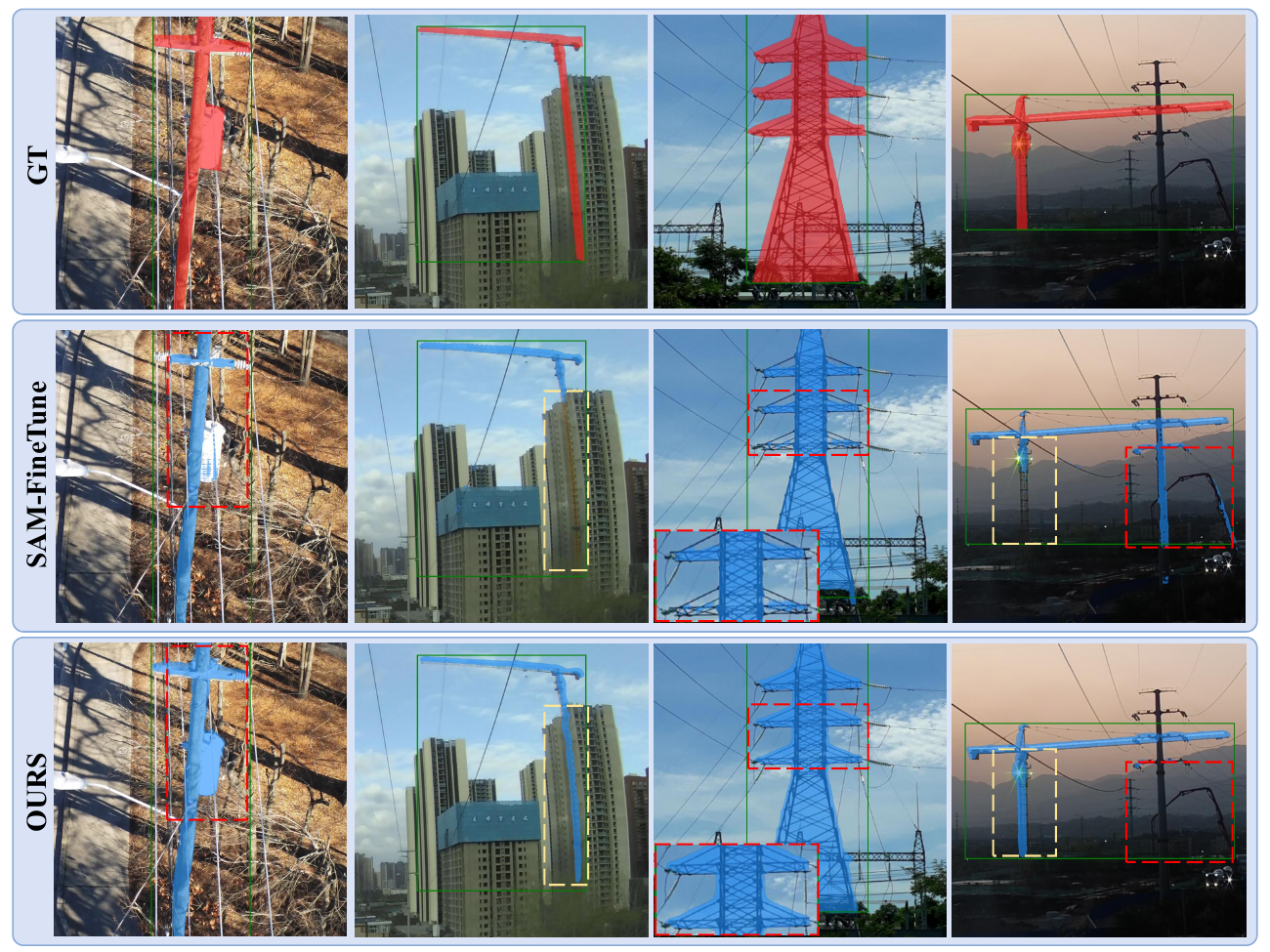}
    \caption{\textbf{Segmentation comparison of SAM and our method in power transmission corridor.} The ground-truth are presented in the first row. Even after fine-tuning, SAM still falls short in segmenting fine-grained objects in complex background.}
\label{fig:fig1}
\vspace{-3mm}
\end{figure}
Recently, for general image segmentation, the Segment Anything Model (SAM)~\citep{kirillov2023segment} has emerged as a foundational model. Benefiting from large-scale pretraining, SAM demonstrates impressive transferability and adaptation across numerous tasks in diverse scenarios, including matting~\citep{li2024matting}, medical image segmentation~\citep{cheng2023sam}, and hierarchical text segmentation~\citep{hisam}. Motivated by the excellent properties of SAM, we leverage it to advance the PTCHS task. However, there are two challenges. 
\textbf{1) Data scarcity.} There are few PTCHS dataset available, lacking adequate hazard categories and high-quality mask annotations in complex transmission corridor scenarios.
\textbf{2) Segmentation quality.} Even after dedicated fine-tuning on PTCHS data, SAM struggles to deal with target objects with fine structure in complex background. For instance, as shown in the second and fourth columns of Fig.~\ref{fig:fig1}, the segmentation on tower crane from SAM is severely interfered by surrounding buildings or equipment.

To solve the above challenges, we start by constructing the first large-scale benchmark for PTCHS, \textbf{ELE-40K}, containing 44,094 pixel-level annotated image-mask pairs derived from real-world transmission corridors. ELE-40K covers transmission equipment, construction vehicles, and environmental hazards, such as wildfire, smoke, \textit{etc}. To obtain the mask annotations, we adopt a semi-automatic and iterative strategy (Kirillov et al., 2023). We first fine-tune SAM with some manual labeled data, and then annotate the remaining images incorporating the iteratively retrained SAM with manual rectification. For PTCHS on ELE-40K, we propose a baseline model named \textbf{ELE-SAM}. Technically, we introduce an Context-Aware Prompt Adapter to improve the prompt tokens. Besides, to better deal with the hazard objects with fine structure during mask decoding, we design a High-Fidelity Mask Decoder. In the mask decoder, multi-sourced features are fused, iteratively refined, and then scaled to a higher resolution of \(512\times 512\). Experimental results demonstrate the effectiveness of our method on the hard samples which require high-quality segmentation capability.

To summarize, our major contributions are three-fold:
\begin{itemize}
\item We construct the ELE-40K dataset, a large-scale benchmark with 44,094 annotated image-mask pairs for real-world power transmission corridor hazard segmentation.
\item We propose the ELE-SAM model, advancing SAM for the PTCHS task. We design two customized modules, the Context-Aware Prompt Adapter for more distinctive prompt tokens, and the High-Fidelity Mask Decoder for segmentation with high-fidelity details.
\item Extensive experiments demonstrate the superior performance of ELE-SAM. Moreover, we also validate the effectiveness of our method on high-quality generic object segmentation, with average 2.9\% mIoU and 3.8\% mBIoU improvements over the leading method.
\end{itemize}

The paper is organized as follows: a brief review of related works is presented in Sec.~\ref{sec: related work}. The proposed method and dataset are illustrated in Sec.~\ref{sec: method}. Extensive experiments are reported in Sec.~\ref{sec: exp}. Some limitations are discussed in Sec.~\ref{sec: limitation}. The paper finally concludes in Sec.~\ref{sec: conclusion}.
\begin{table*}[htbp]
    \centering
    \setlength{\tabcolsep}{8pt}
    \caption{
    \textbf{Comparison of related datasets in terms of image quantity, annotation type, public availability, target.} Our ELE-40K provides the most extensive annotations, focusing on diverse equipment and hazards.
    }
    \label{tab:related-datasets}
    \small 
    \resizebox{1\hsize}{!}{
    \begin{tabular}{lccccc}
         \toprule[1.1pt]
          & &Annotation & Public & \multicolumn{2}{c}{Targets} \\
         \multirow{-2}{*}{Datasets} &\multirow{-2}{*}{Total Images} &Format &Availability &Equipment & Hazards \\
         \midrule
         Carlos \etal~\citep{6889836} &3,200 &Bounding Box &No &1 &-  \\
         NAL-RGB~\citep{inbook} &3,568 &Binary Mask  &No &1 &-  \\
         PLDU~\citep{rs11111342} &573 &Binary Mask &Yes &1 &-  \\
         PLDM~\citep{rs11111342} &287 &Binary Mask &Yes &1 &-  \\
         Vepl~\citep{data8080128} &3,724 & Binary Mask &Yes &2 &1  \\
         DS1\_Co~\citep{8610301} &28,674 & Bounding Box &No &1 &-  \\
         SR-RGB~\citep{Yetgin2017PowerlineID} &2,000 &Class Label &Yes &1 &-  \\
        TTPLA~\citep{abdelfattah2020ttplaaerialimagedatasetdetection} &1,100 &Binary Mask &Yes &4 &-  \\
         \rowcolor{gray!10}ELE-40K &44,094 &Binary Mask &Yes &4 &11  \\
         \bottomrule[1pt]
    \end{tabular}}
    \vspace{-2mm}
\end{table*}

\section{Related Work}\label{sec2}

\label{sec: related work}
\subsection{Power Transmission Corridor Inspection}
Recently, the application of computer vision technology in power transmission corridor inspection arouses increasing attention~\citep{10445528,Nair2024}. Object detection methods are primarily adopted. These methods are categorized into two-stage and single-stage algorithms. Two-stage methods extract candidate regions before detection, offering higher accuracy~\citep{9301267,https://doi.org/10.1049/gtd2.12688}, while single-stage methods perform end-to-end detection using convolutional neural networks, prioritizing speed~\citep{9530535,Tang2024}. However, these object detection methods solely predict bounding boxes that include background objects, hindering precise contour delineation of hazardous objects. This limitation impedes morphological analysis and reduces the accuracy of tasks like distance measurement. In contrast, power transmission corridor hazard segmentation (PTCHS) emerges as an optimal approach~\citep{SHEN2023103263,DBLP:journals/corr/RonnebergerFB15,rs16234585}. Wei \etal~\citep{ Wei2022} reduce the cost of manual labeling and improve the performance of power line segmentation based on the Swin-Unet framework with improved linear embedding and efficient sample synthesis techniques. Hu \etal~\citep{rs16234585} introduce a gated axial attention mechanism and a local normalization module for axial channels. Abdelfattah\etal~\citep{10301645} propose a novel framework, which leverages adversarial training, a Hough transform loss function, and a semantic decoder to achieve excellent performance in segmenting power lines. However, these methods are limited to specific transmission equipment or hazards and cannot address dynamic risks in real-world transmission corridors. Thus, a generalized segmentation model with robust adaptability is urgently required to address this challenge.

\subsection{Segment Anything Model}

The Segment Anything Model (SAM)~\citep{kirillov2023segment} advances the segmentation field through large-scale pretraining and enabling interactive visual prompts. The extraordinary performance and favorable generalization across various real-world scenarios render SAM as one of the vision foundation models. However, SAM's zero-shot performance in a few specialized fields exhibits a notable decrease when encountering unseen features~\citep{wang2023sammeetsroboticsurgery,mohapatra2023samvsbetcomparative}. To promote the adaptability, researchers have improved the architecture of SAM to better deal with specialized challenges, such as camouflaged object detect~\citep{chen2024samcodsamguidedunifiedframework} and medical image analysis ~\citep{Ma2024,10656025}. For instance, Chen \etal~\citep{chen2024samcodsamguidedunifiedframework} enhance SAM's prompt adaptation capability by introducing a response filter and semantic matcher, thereby improving mask quality in camouflaged object detection scenarios. HQ-SAM~\citep{sam_hq} enhances SAM’s segmentation precision by integrating a learnable High-Quality Output Token within the mask decoder, leveraging effective fusion of features from ViT layers. PA-SAM~\citep{xie2024pasampromptadaptersam} refines SAM’s segmentation capabilities through a prompt adapter, which optimizes feature extraction and decoding for improved flexibility and performance. Hi-SAM~\citep{hisam} realizes hierarchical text segmentation in a unified framework. In contrast, to overcome the limitations of SAM in the PTCHS task, we introduce ELE-SAM, a tailored framework for PTCHS. ELE-SAM integrates a Context-Aware Prompt Adapter, a High-Fidelity Mask Decoder, and leverages the ELE-40K dataset to address critical challenges, including fine-structure hazard detection, complex background segmentation, and data scarcity.
\begin{figure*}
    \centering
    \includegraphics[width=1\linewidth]{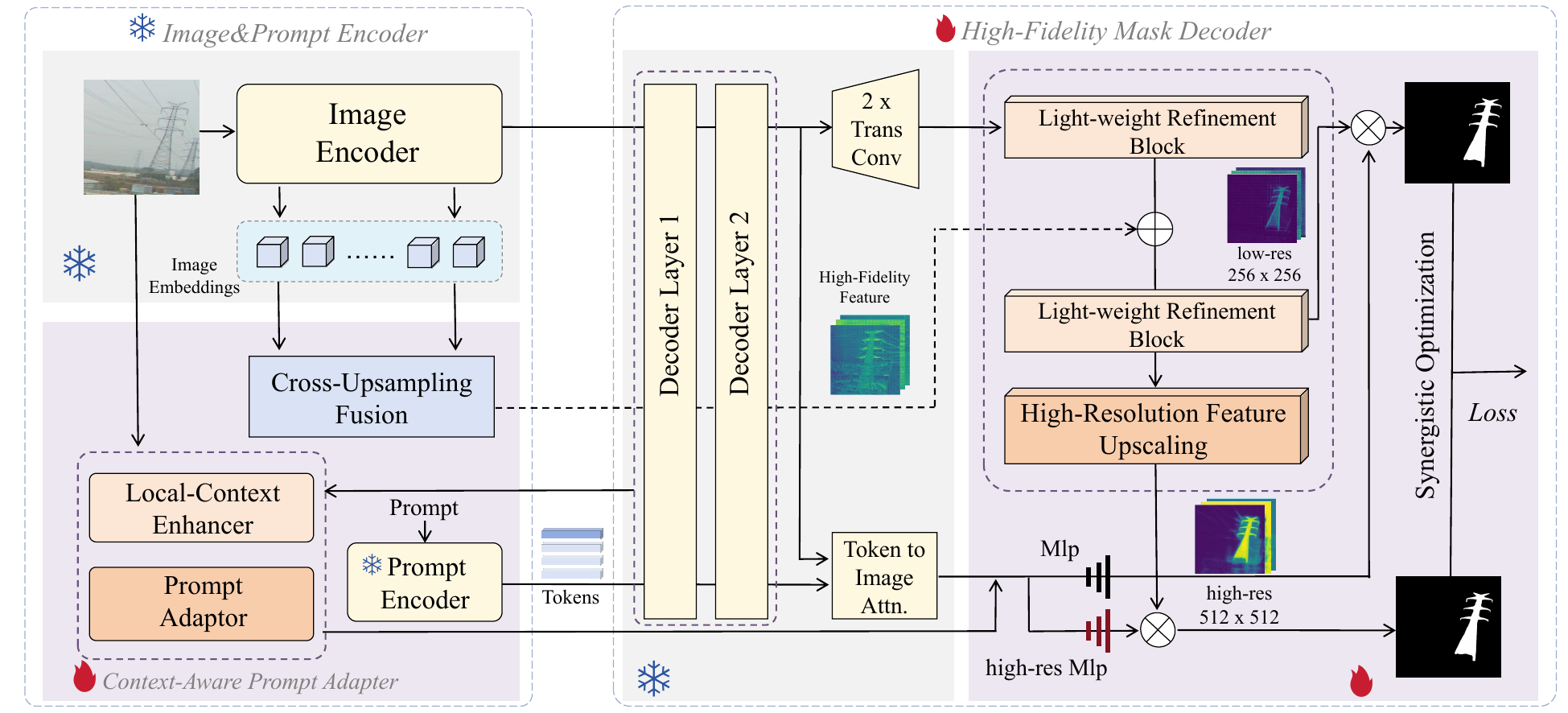}
    \caption{\textbf{The overall architecture of ELE-SAM.} ELE-SAM employs an encoder-decoder paradigm as SAM, incorporating two novel modules, the context-aware prompt adapter (CAPA) and the high-fidelity mask decoder (HFMD). CAPA generates more discriminative prompt tokens to better distinguish the target object. For high-quality segmentation, HFMD further produces mask features in a higher resolution of $512 \times 512$, overcoming the perception loss on object details with the original mask feature resolution.}
    \label{fig:overall}
\end{figure*}
\subsection{Benchmark Datasets} Publicly available datasets for PTCHS are scarce, limiting the development and evaluation of advanced detection models. General datasets like COCO \citep{singh2024benchmarkingobjectdetectorscoco} and ImageNet \citep{5206848} lack power-specific annotations, focusing on generic objects rather than the structurally complex hazards and transmission equipment. 
While some small-scale datasets capture power transmission scenes via drone or high-resolution imagery, their limited scope and variety hinder model generalization across diverse settings \citep{data8080128,abdelfattah2020ttplaaerialimagedatasetdetection,8610301}. For instance, as presented in Tab.~\ref{tab:related-datasets}, although the TTPLA~\citep{abdelfattah2020ttplaaerialimagedatasetdetection} introduces mask annotations, it primarily focuses on a limited range of transmission equipment, such as utility poles and lines, while critical components like transmission towers remain sparsely annotated. Furthermore, the dataset lacks attention to potential hazards in the surrounding environment that could pose threats to transmission equipment. These limitations collectively render existing datasets insufficient for providing comprehensive training in the context of power transmission corridor inspection. To this end, we introduce the ELE-40K dataset, designed to include diverse equipment and hazard scenarios, providing a robust benchmark to improve model accuracy and reliability in real-world PTCHS. 

\section{Methodology}
\label{sec: method}
In this work, we propose \textbf{ELE-SAM} and \textbf{ELE-40K} benchmark. We firstly offer an overview of ELE-SAM in Sec.~\ref{subsec: Overview of ELE-SAM}. Then, we provide detailed method description in subsequent subsections. We also describe the construction procedure and statistic of ELE-40K in Sec.~\ref{subsec: ELE-40K Benchmark Dataset}.

\subsection{Overview of ELE-SAM}
\label{subsec: Overview of ELE-SAM}
As depicted in Fig.~\ref{fig:overall}, ELE-SAM consists of four major components: 1) a frozen image encoder from SAM~\citep{kirillov2023segment}, 2) a frozen SAM's prompt encoder for encoding the initial prompt tokens, 3) a plug-and-play \textbf{Context-Aware Prompt Adapter} (CAPA) for mining more discriminative prompt tokens, and 4) a customized \textbf{High-Fidelity Mask Decoder} (HFMD) for producing and refining mask features in bi-resolution. 

Concretely, given the input image $\bm{I}$, the image encoder generates image embedding $\bm{I}_{emb}$. Following~\citep{sam_hq}, we also extract and fuse the early layer and final layer features from image encoder, resulting in the fused features $\bm{I}_{fusion}$. Note that the image encoder is the same as that in SAM, without any learnable component inserted. 
Meanwhile, visual prompts like boxes are embedded by the prompt encoder and combined with output token, forming the initial prompt tokens $\bm{P}$. 

In HFMD, the two-way Transformer decoder layers firstly interact with CAPA to generates enhanced prompt tokens, including quantity-augmented sparse prompt tokens. Specifically, with the image $\bm{I}$, image embedding $\bm{I}_{emb}$, and prompt tokens $\bm{P}$, after the two-way decoder layers that interact with CAPA, promoted image embedding $\bm{I}_{emb}'$ and enhanced prompt tokens $\bm{P}'$ are obtained.
Then, $\bm{I}_{emb}'$ is upsampled to $256 \times 256$ in resolution, forming the mask features $\bm{F}$. After the final token-to-image attention, the output token $\bm{T}$ is separated from $\bm{P}'$ for subsequent mask prediction.
Finally, in our newly introduced modules for producing and refining mask features in bi-resolution, given fused features $\bm{I}_{fusion}$, mask features $\bm{F}$, and the output token $\bm{T}$, mask logits with $256 \times 256$ and $512 \times 512$ resolution can be achieved respectively. 
In the following subsections, we delve into the technical details of CAPA and HFMD.

\subsection{Context-Aware Prompt Adapter}
\begin{figure}[!t]
    \centering
\includegraphics[width=0.8\linewidth]{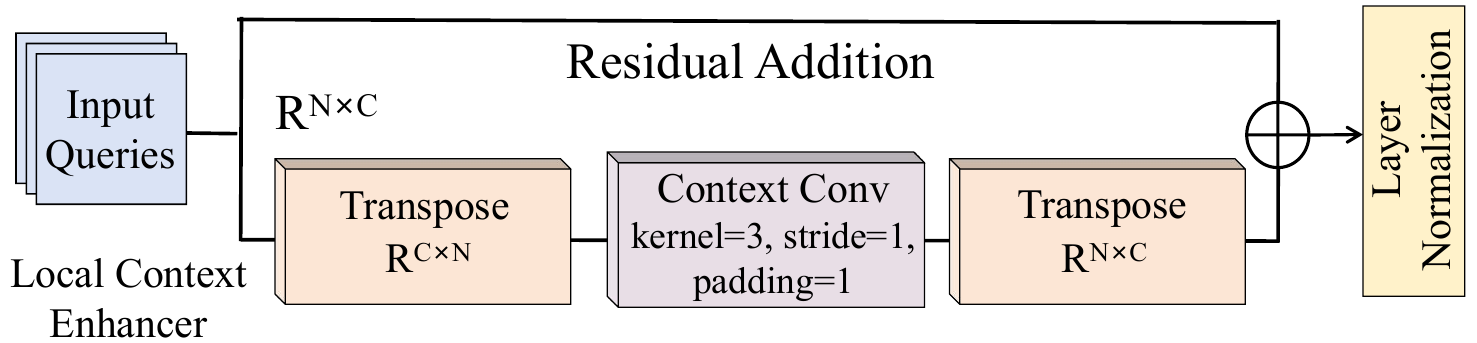}
    \caption{\textbf{The structure details of Local Context Enhancer.}}
    \label{fig: LCE}
\end{figure}

Given the initial prompt tokens $\bm{P}$, $\bm{P}$ could be coarse and inadequate to determine some uncertain regions in high-quality segmentation.
Inspired by PA-SAM~\citep{xie2024pasampromptadaptersam}, we incorporate its Prompt Adapter (PA) to generate more distinctive prompt tokens. 
Differently, we additionally design a Local Context Enhancer (LCE) and insert it before PA to enable pre-interaction between prompts, thereby enhancing the adaptive prompt generation. The structure of LCE is illustrated in Fig.~\ref{fig: LCE}.
Specifically, given the prompt tokens $\bm{P} \in \mathbb{R}^{N \times C}$, where $N$ is token number and $C$ is dimension, LCE conducts pre-interaction among prompts with a simple convolutional operation and residual addition:
\begin{equation}
    \hat{\bm{P}} = LN(Conv(\bm{P}) + \bm{P}),
\end{equation}
where $LN$ represents layer normalization and $Conv$ stands for 1D convolutional operation (kernel size: 3, stride: 1, padding: 1). $\hat{\bm{P}}$ represents the obtained intermediate prompt tokens. The promoted image embedding $\bm{I}_{emb}'$ and enhanced prompt tokens $\bm{P}'$ can be achieved with the prompt adapter $PA$ as follows:
\begin{equation}
    \bm{I}_{emb}', \bm{P}' = PA(\bm{I}, \bm{I}_{emb}, \hat{\bm{P}}),
\end{equation}
where $\bm{I}_{emb}' \in \mathbb{R}^{64 \times 64 \times 256}$ keeps the same shape as $\bm{I}_{emb}$, enhanced prompt tokens $\bm{P}' \in \mathbb{R}^{M \times C}$ are augmented in token number with more distinctive sparse prompts.

The prompt adapter $PA$ consists of two key components: 1) \textbf{Adaptive Detail Enhancement}. This scheme explores detail information from the image and its Canny gradient by Dense Prompt Compensation and Sparse Prompt Optimization. 2) \textbf{Hard Point Mining}. This operation samples more positive and negative points, then embeds and concatenates them to the input prompts $\hat{\bm{P}}$. In the prompt adapter, coarse, refined, and uncertain masks are produced and used to calculate the loss $\mathcal{L}_{\text{PA}}$. We follow the same implementation as in PA-SAM~\citep{xie2024pasampromptadaptersam}, where more details are expanded.

\begin{figure}[!t]
    \centering
\includegraphics[width=0.8\linewidth]{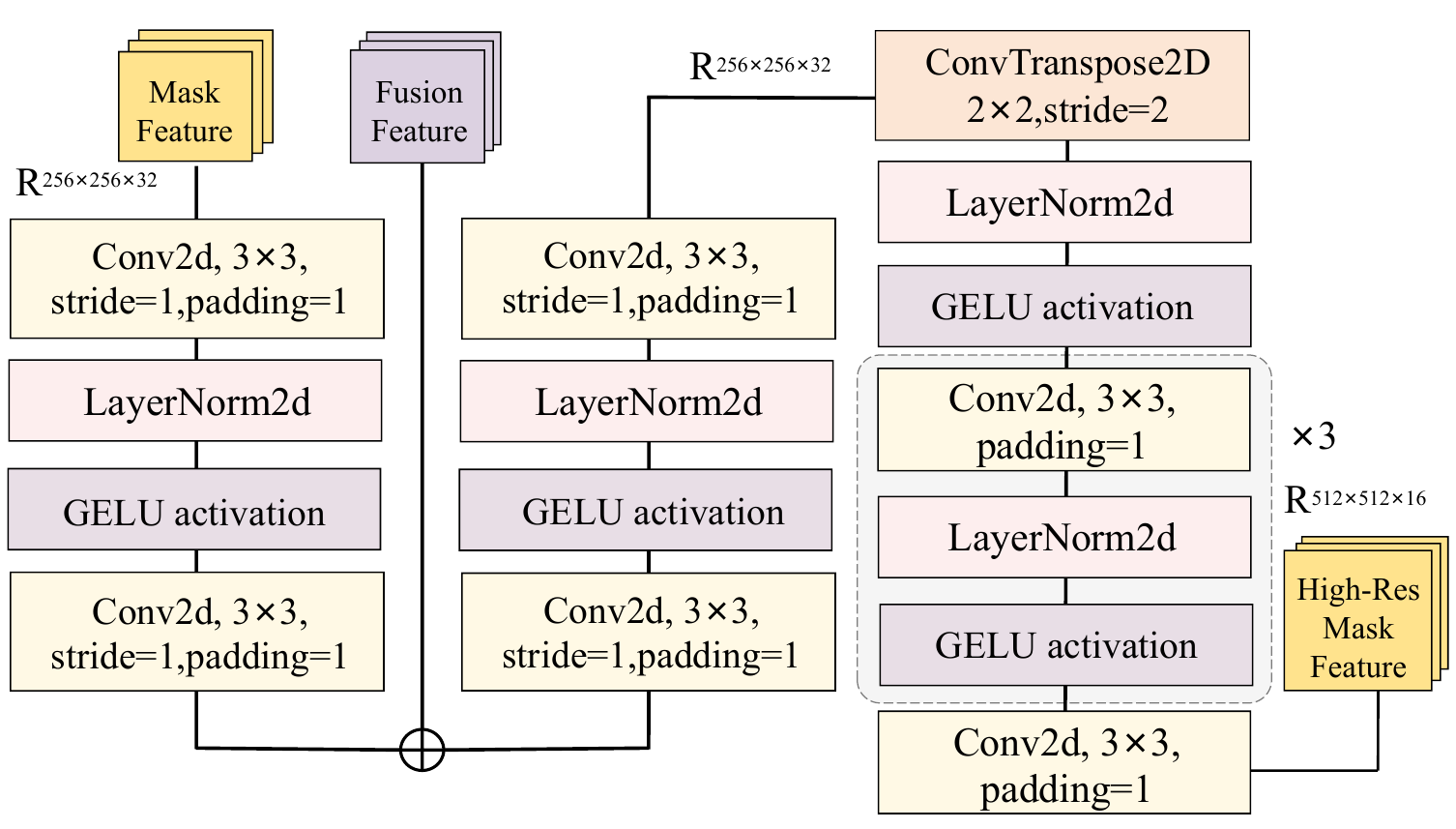}
    \caption{\textbf{The structural details for producing and refining the mask features in low-resolution of $256 \times 256$ and high-resolution of $512 \times 512$.}}
    \label{fig: HFD}
\end{figure}

\subsection{High-Fidelity Mask Decoder}
With the promoted image embedding $\bm{I}_{emb}'$ and enhanced prompt tokens $\bm{P}'$, mask features $\bm{F} \in \mathbb{R}^{256 \times 256 \times 32}$ are obtained by applying two transposed convolution layers on $\bm{I}_{emb}'$. After the final token-to-image attention, the output token $\bm{T} \in \mathbb{R}^{1 \times 256}$ is sliced from $\bm{P}'$ for mask prediction.

Then, the output token $\bm{T}$, mask features $\bm{F}$, and fused features $\bm{I}_{fusion}$ are send into our newly designed components for producing and refining mask features in bi-resolution, as shown in Fig.~\ref{fig: HFD}. Concretely, the mask features $\bm{F}$ are processed with convolution and added on $\bm{I}_{fusion}$:
\begin{equation}
    \bm{F}_{fusion} = Conv2DBlock_1(\bm{F}) + \bm{I}_{fusion},
\end{equation}
where $Conv2DBlock_1$ consists of two 2D convolutional layers, with 2D layer normalization and GELU activation inserted between them.

Then, the features $\bm{F}_{fusion}$ are further refined with a convolutional block which shares the same parameters with $Conv2DBlock_1$:
\begin{equation}
    \bm{F}_{fusion}' = Conv2DBlock_1(\bm{F}_{fusion}).
\end{equation}

In this way, the low-resolution mask prediction $\bm{M}_{\text{lr}}$ can be achieved based on $\bm{F}_{fusion}' \in \mathbb{R}^{256 \times 256 \times 32}$ and $\bm{T}$:
\begin{equation}
    \bm{M}_{\text{lr}} = \bm{F}_{fusion}' \odot MLP_{\text{lr}}(\bm{T}),
\end{equation}
where $\odot$ represents the dot-product operation, $MLP_{\text{lr}}$ is a three-layer multi-layer perceptron (MLP) which projects $\bm{T}$ and reduces its dimension to $32$.

To achieve high-resolution mask features $\bm{F}_{\text{hr}} \in \mathbb{R}^{512 \times 512 \times 16}$, where $16$ is the feature dimension, we further upsample $\bm{F}_{fusion}'$ with a transposed convolution $TransConv2D$ and employ a block $Conv2DBlock_2$ with four convolutional layers for refinement:
\begin{equation}
    \bm{F}_{\text{hr}} = Conv2DBlock_2(TransConv2D(\bm{F}_{fusion}')).
\end{equation}

The transposed convolution upsamples the resolution of $\bm{F}_{fusion}'$ to $512 \times 512$ while reducing the dimension to $16$. Layer normalization and GELU activation are also inserted, as illustrated in Fig.~\ref{fig: HFD}.

Finally, given the high-resolution mask features $\bm{F}_{\text{hr}}$ and output token $\bm{T}$, the mask prediction $\bm{M}_{\text{hr}}$ in high-resolution of $512 \times 512$ can be obtained:
\begin{equation}
    \bm{M}_{\text{hr}} = \bm{F}_{\text{hr}} \odot MLP_{\text{hr}}(\bm{T}),
\end{equation}
where $MLP_{\text{hr}}$ is also a three-layer MLP. $MLP_{\text{hr}}$ projects $\bm{T}$ and reduces its dimension to $16$.
\begin{figure*}
    \centering
    \includegraphics[width=1\linewidth]{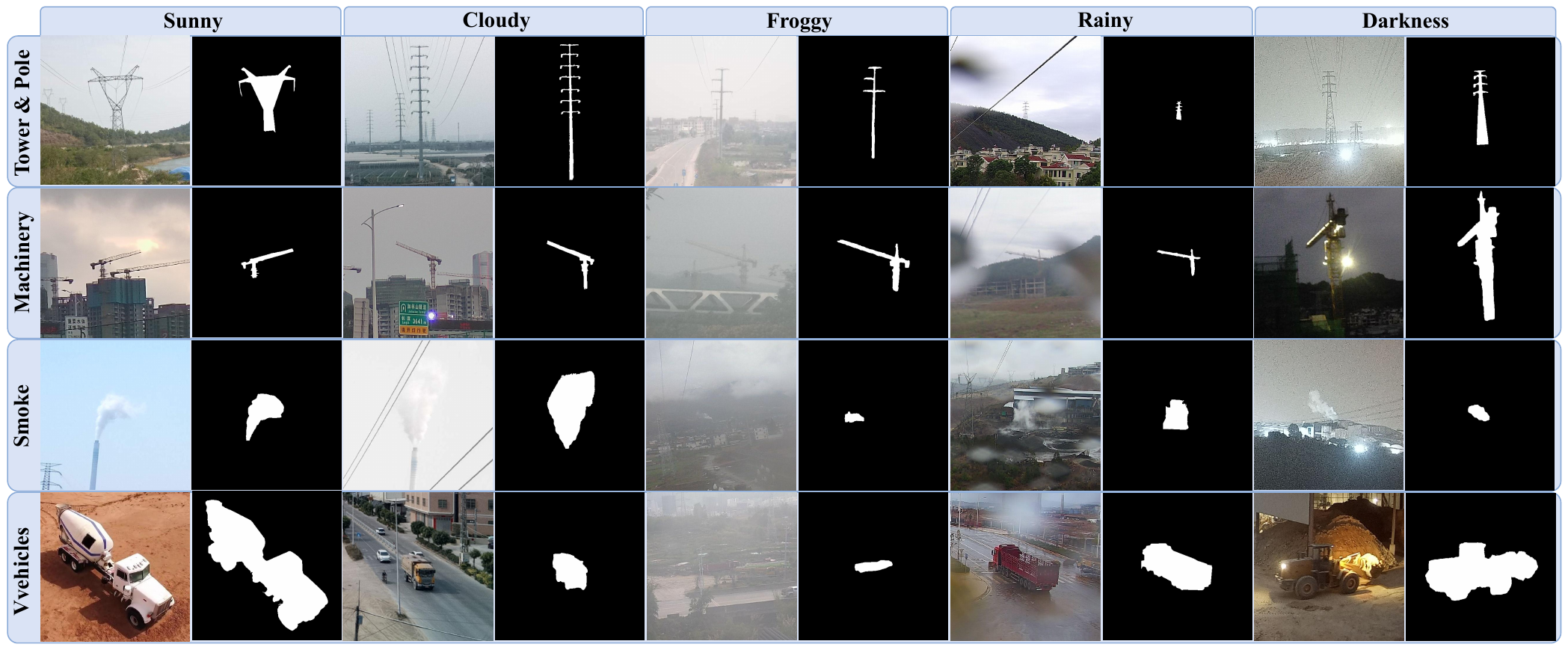}
    \caption{\textbf{Samples in ELE-40K under different imaging conditions.} It shows transmission equipment and surrounding hazards under various conditions such as sunny, overcast, fog, rain, and darkness. This highlights the comprehensiveness of the dataset.}
    \label{eledataset}
\end{figure*}

\begin{figure}[!t]
    \centering
\includegraphics[width=0.7\linewidth]{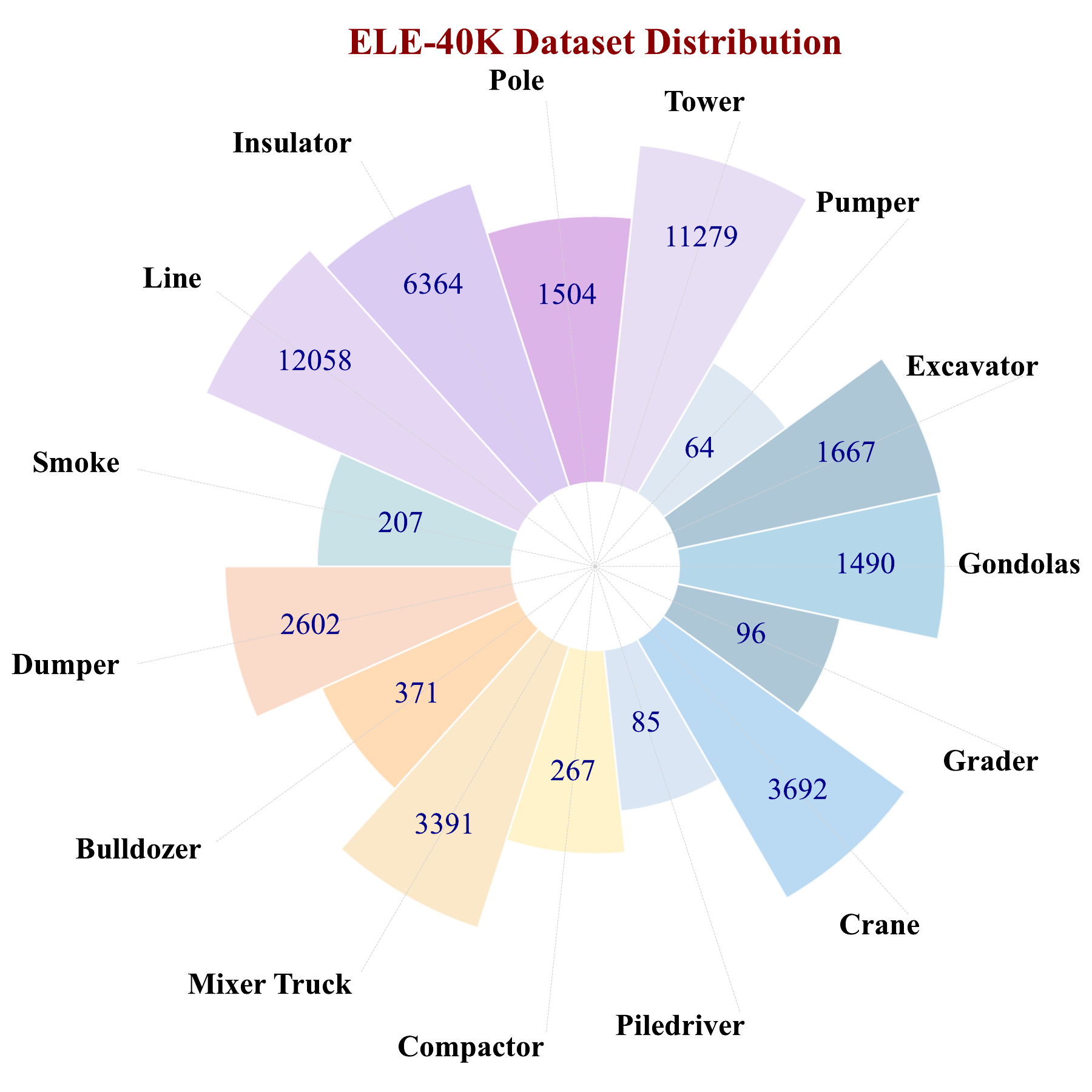}
    \caption{\label{fenbu} The data distribution of ELE-40K, which includes 31,205 annotated transmission equipment instances and 12,889 annotated surrounding hazard instances. }
\end{figure}

\begin{figure*}[htbp] 
\captionsetup[subfigure]{labelfont=rm,font=rm}
\renewcommand{\thesubfigure}{(\alph{subfigure})}
\centering
\subfloat[]{\includegraphics[width=\textwidth]{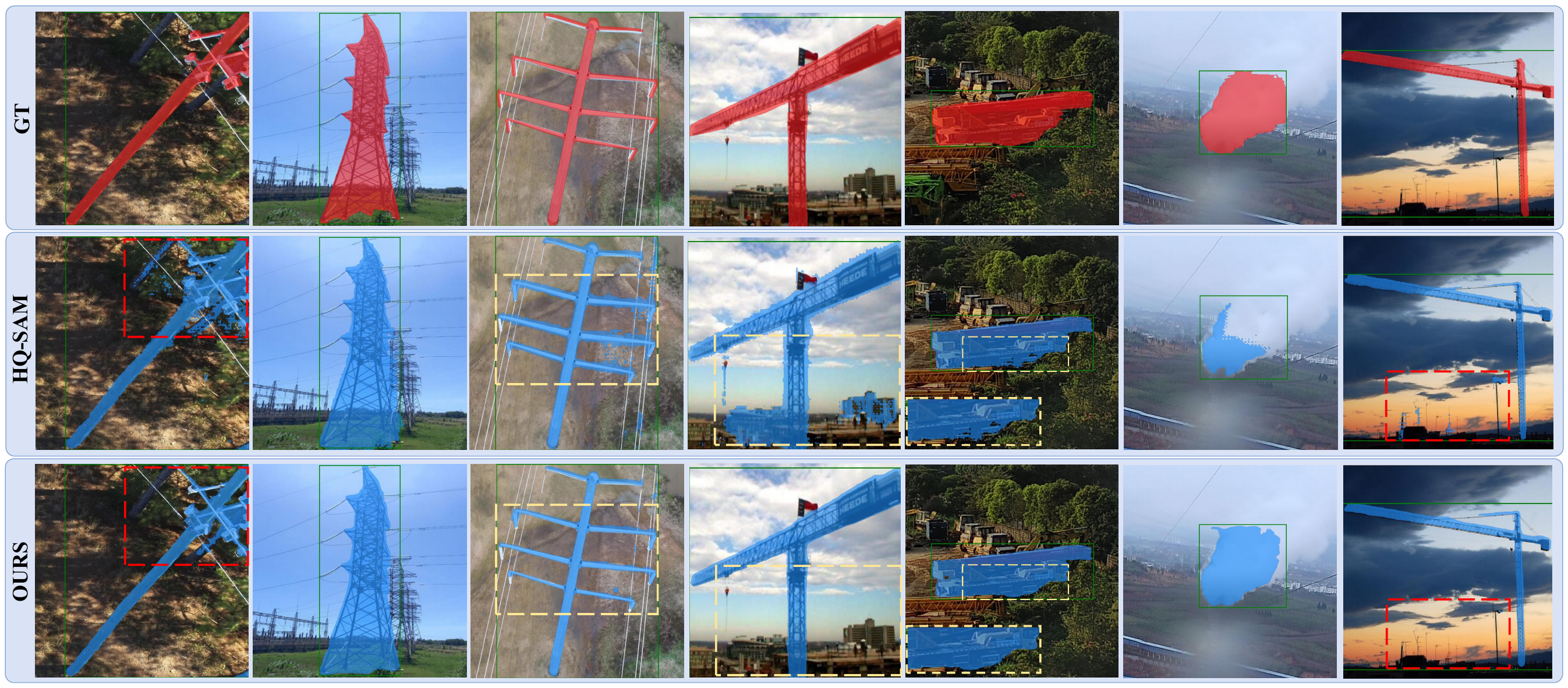}%
\label{fig:7a}} \\
\vspace{-0.4cm} 
\subfloat[]{\includegraphics[width=\textwidth]{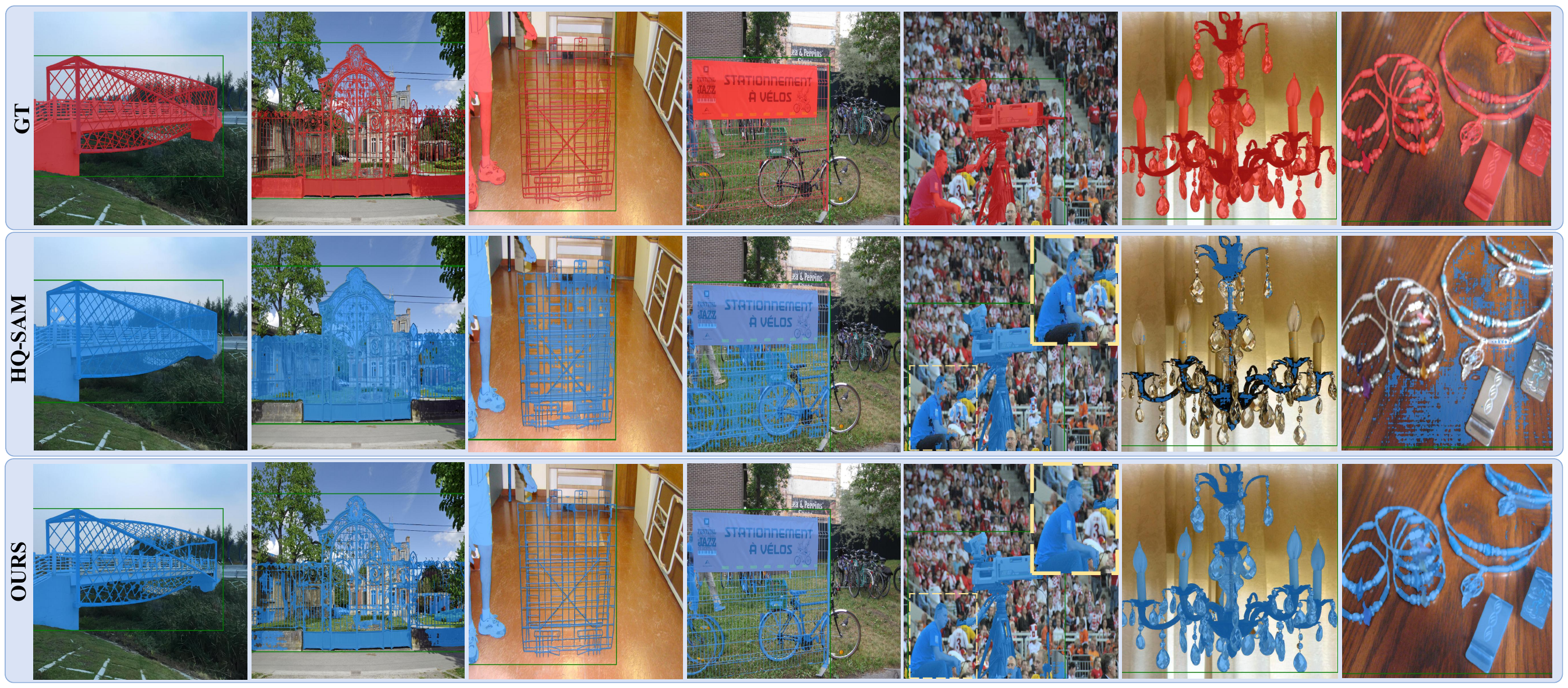}%
\label{fig:7b}}
\caption{\rmfamily From the first to the third row, we present the ground truth, the suboptimal method (HQ-SAM~\citep{sam_hq}), and our ELE-SAM, respectively. Specifically, (a) illustrates the visualization results on ELE-40K dataset, while (b) shows the results on HQSeg44K dataset.}
\end{figure*}

\subsection{Loss Function}

Since ELE-SAM predicts bi-resolution results, we apply supervision for both of them to ensure coarse-to-fine mask evolution. In particular, given low-resolution prediction $\bm{M}_{\text{lr}}$ and high-resolution prediction $\bm{M}_{\text{hr}}$, we calculate the Dice loss~\citep{dice} for each resolution level with ground-truth $\bm{M}_{\text{GT}}$. The total loss $\mathcal{L}$ equals to the Dice loss from two resolution levels plus the loss $\mathcal{L}_{\text{PA}}$ from prompt adapter, which is formulated as:

\begin{align}
\mathcal{L} =  \mathcal{L}_{\text{Dice}}(\bm{M}_{\text{lr}}, \bm{M}_{\text{GT}}) + \mathcal{L}_{\text{Dice}}(\bm{M}_{\text{hr}}, \bm{M}_{\text{GT}}) + \mathcal{L}_{\text{PA}}.
\end{align}

\subsection{ELE-40K Benchmark}
\label{subsec: ELE-40K Benchmark Dataset}
The increasing demand for electric power has highlighted the importance for monitoring the power transmission corridors and maintaining transmission safety. Progress in PTCHS remains limited due to the absence of large-scale and high-quality dataset that features real-world complexity. Thus, we construct ELE-40K, a benchmark dataset with 44,094 pixel-annotated image-mask pairs from real transmission systems, offering a solid foundation for advancing PTCHS.

\noindent\textbf{Dataset Sources.} ELE-40K dataset is derived from two primary sources:
1) Self-collected dataset: We collected and annotated a total of 22,878 image-mask pairs captured by unmanned aerial vehicles and cameras along transmission corridors. 2) Publicly available datasets: We extracted and re-annotated data from existing publicly available datasets ~\citep{abdelfattah2020ttplaaerialimagedatasetdetection, towerdetection_dataset, electricpolefull-yxgbo_dataset, cranes-xxz5d_dataset, excavators-cwlh0_dataset}, covering instances such as power lines, insulators, poles and engineering vehicles. Combining these two sources, we obtained a total of 31,205 annotated transmission equipment and 12,889 hazards image-mask pairs.

\noindent\textbf{Dataset Construction and Statistic.}
The images in ELE-40K are from real-world transmission corridor scenario, capturing diverse operational and environmental conditions. To achieve segmentation annotations, inspired by SAM (Kirillov et al., 2023), we employ a semi-automated annotation process. The process includes: 1) Manual initial annotating. Domain experts annotate images at the pixel level, identifying transmission equipment and potential hazards. In the initial stage, we annotated 1,000 image-mask pairs.
2) We train ELE-SAM with pixel-level annotation from step 1. The derived model is then applied to predict initial masks for a new set of 2,000 images, accelerating the annotation process.
3) Experts manually refine the false positive and false negative regions in the preliminary masks generated in step 2 to ensure annotation accuracy. 
4) We combine the data in step 1 and refined data for retraining ELE-SAM to improve accuracy in subsequent iterations. The loop of SAM-assisted labeling, manual refinement, and retraining is repeated until all data is fully annotated. 

Due to annotation cost and the practical requirement that subsequent tasks like distance measurement and risk assessment rely on overall object contours, the hollows between steel wires inside the objects have not been accurately annotated. During evaluation, we follow the promptable segmentation paradigm as in HQSeg-44K, where each object is individually segmented based on the given bounding box prompt. As modern detectors can provide high-accuracy boxes in real-time, we focus on category-agnostic promptable segmentation.
Finally, the dataset contains key transmission equipment such as transmission towers, poles, and support structures. Frequent hazards encountered during maintenance activities are also included, such as engineering vehicles, bulldozers, cranes, and other heavy machinery. To further enhance applicability, high-impact environmental hazards, such as wildfire and smoke, are also included. ELE-40K features different imaging condition, including froggy, rainy, and darkness. Some examples are shown in Fig.~\ref{eledataset}. The detailed instance distribution is illustrated in Fig.~\ref{fenbu}.
\begin{table*}[htbp]
\centering
\caption{\textbf{Performance on ELE-40K}, including detailed results for equipment and hazards. SAM-FineTune
denotes fine-tuning the entire mask decoder of SAM}
\setlength{\tabcolsep}{14pt}
\resizebox{\linewidth}{!}{
\begin{tabular}{l| c c | c c| c c}
\toprule
\multirow{2}{*}{Model} & \multicolumn{2}{c|}{Equipment} & \multicolumn{2}{c|}{Hazards}  & \multicolumn{2}{c}{Average} \\
\cmidrule(lr){2-3} \cmidrule(lr){4-5} \cmidrule(lr){6-7}
 & mIoU & mBIoU & mIoU & mBIoU & mIoU & mBIoU \\
\midrule

SAM~\citep{kirillov2023segment}  & 52.3 & 42.5 & 74.1 & 64.6 & 63.2 & 53.5        \\

SAM-FineTune & 55.1 & 45.2 & 75.6 & 67.8 & 65.4 & 56.5         \\
U²Net~\citep{fang2021denselynestedtopdownflows}    & 41.8 & 38.1 & 56.8 & 49.7 & 49.3 & 43.9         \\
IS-Net-General-Use~\citep{qin2022highlyaccuratedichotomousimage}      & 26.5 & 19.9 & 14.0 & 11.2 & 20.3 & 15.6         \\
IS-Net      & 47.9 & 44.6 & 62.9 & 55.4 & 55.4 & 50.0         \\
MvaNet~\citep{yu2024multiview}      & 43.6 & 39.2 & 58.0 & 45.7 & 50.1 & 44.1         \\
HQ-SAM~\citep{sam_hq}   & 44.8 & 29.5 & 70.7 & 63.6 & 57.7 & 46.5        \\
HQ-SAM-FineTune          & 69.7 & 66.4 & 79.4 & 73.3 & 74.6 & 69.8         \\
PA-SAM~\citep{xie2024pasampromptadaptersam}          & 36.3 & 29.6 & 72.2 & 64.3 & 54.3 & 47.0 \\
PA-SAM-FineTune          & 71.3 & 65.2 & 80.8 & 71.9 & 76.0 & 68.6 \\
\rowcolor{gray!10}ELE-SAM & \textbf{76.3} & \textbf{73.4} & \textbf{83.8} & \textbf{74.9} & \textbf{80.0} & \textbf{74.1} \\
\bottomrule
\end{tabular}
\label{tab:comparison_results_adjusted}
 }
\end{table*}
\section{Experiments}
\label{sec: exp}
\subsection{Experiment Details}
\textbf{Datasets.} \textbf{ELE-40K} comprises 44,094 image-mask pairs annotated with real-world transmission equipment and hazard scenarios, with 80\% of the data allocated for training and 20\% for validation. \textbf{HQSeg-44K}~\citep{sam_hq} is a comprehensive dataset combining six existing image segmentation datasets, including the training sets of DIS~\citep{qin2022highlyaccuratedichotomousimage}, ThinObject-5K~\citep{9423134}, FSS~\citep{Li_2020}, ECSSD~\citep{7182346}, MSRA-10K~\citep{6871397}, and DUT-OMRON~\citep{6619251}. The DIS, ThinObject-5K, COIFT, and HR-SOD datasets are used as validation sets. This results in a total of 44,320 annotated image-mask pairs. For the division of the dataset, we adopt the same configuration as HQ-SAM and PA-SAM. Additionally, we use \textbf{COCO}\citep{singh2024benchmarkingobjectdetectorscoco} for evaluating zero-shot segmentation.

\textbf{Evaluation Metrics.} Following HQ-SAM~\citep{sam_hq} for assessing high-quality segmentation, we adopt mean Intersection over Union (mIoU) and mean Boundary Intersection over Union (mBIoU) as metrics. 

\textbf{Implementation Details.} 
For fair comparison, we keep the training configurations of ELE-SAM consistent with other SAM-based models. Specifically, a learning rate of \(1 \times 10^{-3}\) is employed for the initial 20 epochs, which is subsequently reduced to \(1 \times 10^{-4}\) for the remaining 10 epochs, resulting in a total of 30 training epochs for both the ELE-40K and HQSeg-44K datasets. We adopt SAM-L as our baseline model, and all experiments related to SAM are conducted using the SAM-L variant to ensure fair comparison. The experiments are conducted using two Nvidia GeForce RTX 3090 (24G) GPUs with the batch size of 8.

\begin{figure}[!t]
    \centering
    \includegraphics[width=1\linewidth]{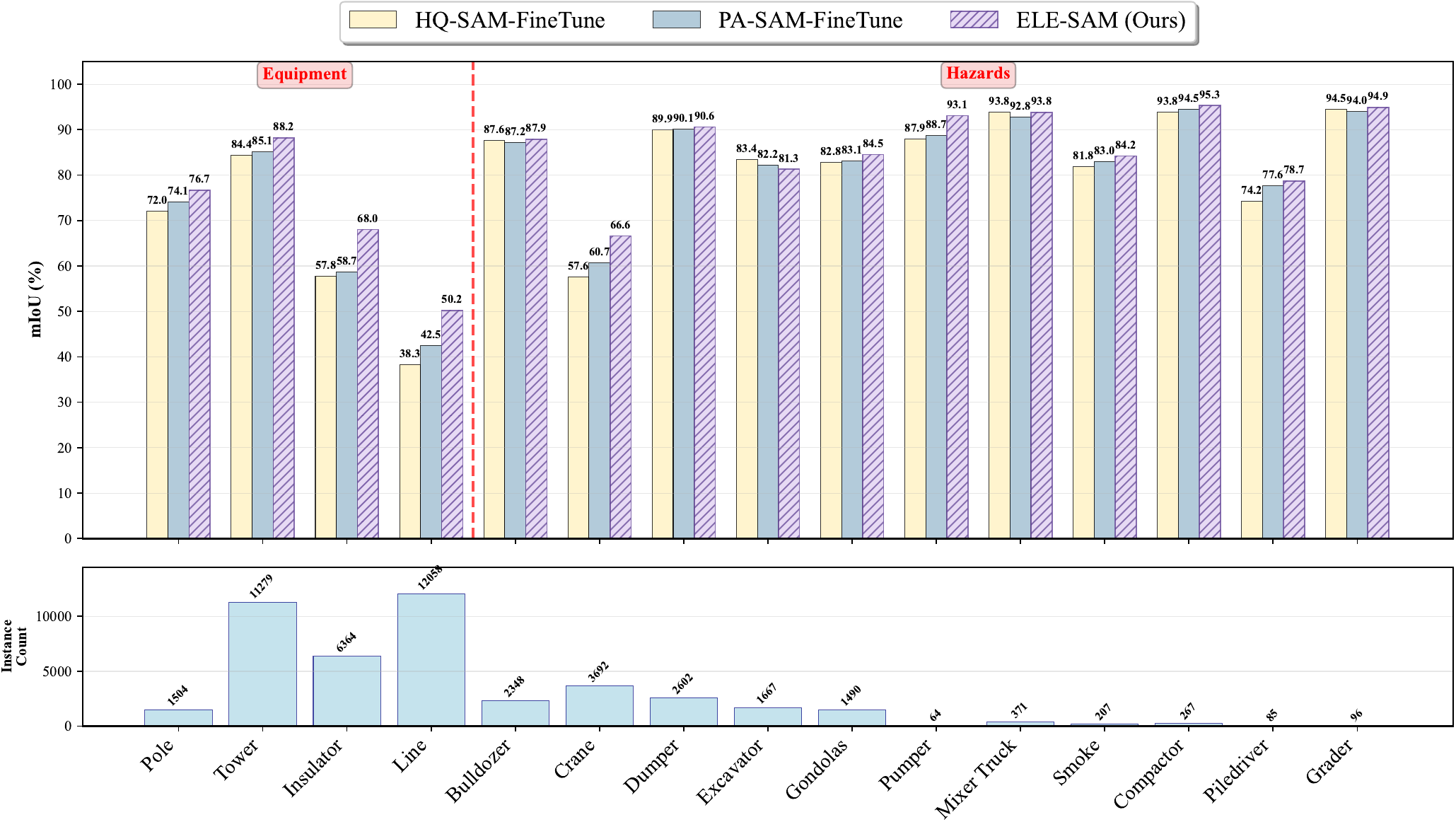}
    \caption{\label{fig:per-category-performance}\textbf{Per-category mIoU comparison on ELE-40K.} 
ELE-SAM achieves superior performance across 14 out of all 15 categories, with significant improvements on Equipment categories featuring fine structures. 
The bottom subplot shows instance counts.}
\end{figure}

\subsection{Results on ELE-40K}
To comprehensively evaluate the effectiveness of ELE-SAM, we conduct comparative experiments on ELE-40K against various state-of-the-art segmentation models, including SAM~\citep{kirillov2023segment}, HQ-SAM~\citep{sam_hq}, PA-SAM~\citep{xie2024pasampromptadaptersam}, U²Net~\citep{fang2021denselynestedtopdownflows}, MvaNet~\citep{yu2024multiview}, and IS-Net~\citep{qin2022highlyaccuratedichotomousimage}. In addition to the overall performance, the detailed results on two primary categories, \ie, equipment and hazards, are also reported. The quantitative results in Tab. \ref{tab:comparison_results_adjusted} demonstrate the superior mIoU and mBIoU performance of ELE-SAM. The results are analyzed in two main aspects: the impact of ELE-40K dataset and the performance superiority of ELE-SAM over other models.

\textbf{Impact of ELE-40K Dataset.} As demonstrated in Tab. \ref{tab:comparison_results_adjusted}, the performance of models like HQ-SAM~\citep{sam_hq}, PA-SAM~\citep{xie2024pasampromptadaptersam}, and IS-Net-General-Use~\citep{qin2022highlyaccuratedichotomousimage} when directly applied to ELE-40K without fine-tuning is significantly lower than their counterparts fine-tuned on ELE-40K (HQ-SAM-finetune~\citep{sam_hq}, PA-SAM-finetune, and IS-Net). For instance, HQ-SAM achieves an mIoU of 44.8\% on equipment and 70.7\% on hazards, while its fine-tuned version obtains absolute improvements of 24.9\% and 8.7\%, respectively. Similarly, PA-SAM shows a substantial improvement from 36.3\% to 71.3\% mIoU on equipment after fine-tuning. This stark contrast indicates the distinctive challenge of the PTCHS task compare to general segmentation, thus highlighting the value of ELE-40K.
\begin{table*}[htbp]
\centering
\scriptsize
\caption{\textbf{Performance on HQSeg-44K with detailed results on its four subsets.}}
\renewcommand{\arraystretch}{1.5} 
\resizebox{\textwidth}{!}{
\begin{tabular}{l|c c| c c |c c| c c| c c c}
\toprule
\multirow{2}{*}{Model} & \multicolumn{2}{c|}{DIS} & \multicolumn{2}{c|}{COIFT} & \multicolumn{2}{c|}{HRSOD} & \multicolumn{2}{c|}{ThinObject} & \multicolumn{2}{c}{Average} \\
\cmidrule(lr){2-3} \cmidrule(lr){4-5} \cmidrule(lr){6-7} \cmidrule(lr){8-9} \cmidrule(lr){10-11}
 &  mIoU & mBIoU & mIoU & mBIoU & mIoU & mBIoU & mIoU & mBIoU & mIoU & mBIoU \\
\midrule
SAM~\citep{kirillov2023segment}  & 62.0 & 52.8 & 92.1 & 86.5 & 90.2 & 83.1 & 73.6 & 61.8 & 79.5 & 71.1 \\
SAM-FineTune    & 78.9 & 70.3 & 93.9 & 89.3 & 91.8 & 83.4 & 89.4 & 79.0 & 88.5 & 80.5 \\
RSPrompter~\citep{10409216}      & 77.8 & 69.9 & 94.5 & 88.7 & 92.4 & 86.5 & 90.0 & 79.7 & 88.7 & 81.2 \\
BOFT-SAM~\citep{liu2024parameterefficientorthogonalfinetuningbutterfly}        & 78.2 & 69.7 & 94.9 & 90.5 & 93.1 & 86.0 & 91.7 & 80.1 & 89.5 & 81.6 \\
HQ-SAM~\citep{sam_hq}          & 78.6 & 70.4 & 94.8 & 90.1 & 93.6 & 86.9 & 89.5 & 79.9 & 89.1 & 81.8 \\
PA-SAM~\citep{xie2024pasampromptadaptersam}          & 81.5 & 73.9 & 95.8 & \textbf{92.1} & 94.6 & 88.0 & 92.7 & 84.0 & 91.2 & 84.5 \\
\rowcolor{gray!10}ELE-SAM & \textbf{88.0} & \textbf{78.0} & \textbf{96.5} & 91.7 & \textbf{96.4} & \textbf{93.6} & \textbf{95.3} & \textbf{90.0} & \textbf{94.1} & \textbf{88.3} \\
\bottomrule
\end{tabular}
\label{hq-seg-44k}
 }
\end{table*}
\textbf{Performance of ELE-SAM.} ELE-SAM consistently outperforms SAM and its derivative models, as well as several state-of-the-art segmentation methods. As shown in Tab. \ref{tab:comparison_results_adjusted}, ELE-SAM outperforms the baseline SAM~\citep{kirillov2023segment} and its fine-tuned version (SAM-finetune)~\citep{kirillov2023segment}, increasing the mIoU by 16.8\% and mBIoU by 20.6\% over SAM, and by 14.6\% in mIoU and 17.6\% in mBIoU over SAM-finetune in average. Furthermore, ELE-SAM achieves clear improvements over other prominent SAM-based models, particularly in equipment segmentation, with the mIoU improvement of 6.6\% over HQ-SAM-finetune~\citep{sam_hq} and 5.0\% over PA-SAM-finetune~\citep{xie2024pasampromptadaptersam}. 
Compared to end-to-end segmentation models, ELE-SAM outperforms U²Net by 30.7\% in mIoU and 30.2\% in mBIoU, while surpassing IS-Net by 24.6\% in mIoU and 24.1\% in mBIoU. Similarly, ELE-SAM achieves a 29.9\% and 30.0\% improvement in mIoU and mBIoU over MvaNet. 

\textbf{Per-Category Analysis.} To provide a more fine-grained evaluation, we further examine the per-category mIoU performance in Fig. \ref{fig:per-category-performance}. ELE-SAM achieves particularly large improvements on challenging equipment categories. For example, for insulator and line, ELE-SAM surpasses PA-SAM-FineTune by 9.3\% and 7.7\% mIoU respectively, underscoring the enhanced ability to handle fine-structured objects.

Overall, benefiting from the proposed CAPA and HFMD modules, which effectively preserve object structural details with high fidelity, ELE-SAM achieves outstanding performance on the ELE-40K dataset and significantly outperforms existing competitors.

\textbf{Visual Results.} In addition to the quantitative results, we also provide some visualizations in Fig.~\ref{fig:7a}. Compared to HQ-SAM~\citep{sam_hq}, ELE-SAM generates sharper and more complete segmentations with fewer false positives, particularly in challenging cases with fine structural details. The visualizations indicate that ELE-SAM excels at distinguishing power transmission equipment from complex background while preserving boundary integrity.

\subsection{Effectiveness on High-Quality Segmentation}
We further validate the effectiveness of our methods on high-quality segmentation for more general objects using HQSeg-44K~\citep{sam_hq}, as shown in Tab.~\ref{hq-seg-44k}.
Overall, ELE-SAM improves the average mIoU and mBIoU by 2.9\% and 3.8\% compared to the previous leading method. 
Regarding the four sub-sets, ELE-SAM obtains more significant enhancement on DIS and ThinObject. For instance, ELE-SAM outperforms PA-SAM~\citep{xie2024pasampromptadaptersam} by 6.5\% mIoU and 4.1\% mBIoU on DIS, 2.6\% mIoU and 6.0\% mBIoU on ThinObject. Additionally, ELE-SAM surpasses HQ-SAM~\citep{sam_hq} by 9.4\% mIoU and 7.6\% mBIoU on DIS, 5.8\% mIoU and 10.1\% mBIoU on ThinObject. Since the two sub-sets contain abundant objects in mesh structure, such as steel cable bridge and iron fence, segmenting these objects in high quality requires larger mask feature resolution. HQ-SAM and PA-SAM only adopt the mask features with $256 \times 256$ resolution, resulting in the perception loss of fine structures. 
As shown in Fig.~\ref{fig:7b}, HQ-SAM fails to segment the structure details of foreground objects. In comparison, ELE-SAM explores the generation and refinement of mask features in $512 \times 512$ resolution, contributing to the substantial improvement of segmentation quality.
\begin{figure}[!t]
    \centering
    \includegraphics[width=0.8\linewidth]{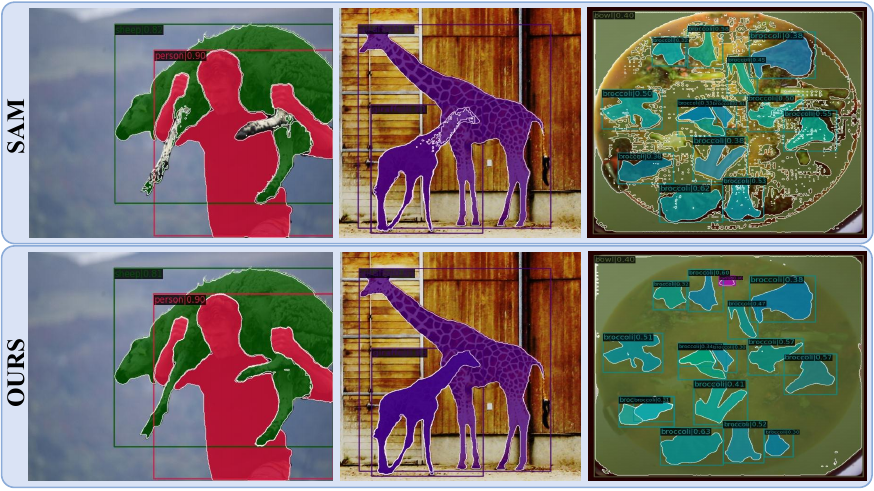}
    \caption{\label{zero-shot-fig}\textbf{Comparison of visual results between SAM (top row) and ELE-SAM (bottom row) on the COCO\citep{singh2024benchmarkingobjectdetectorscoco} validation set under a zero-shot setting.} FocalNet-DINO~\citep{zhang2022dinodetrimproveddenoising}, trained on the COCO dataset is utilized as the box prompt generator. ELE-SAM demonstrates superior mask quality compared to SAM, achieving high-accuracy segmentation while maintaining robust zero-shot segmentation performance.}
\end{figure}
\begin{figure}[!t]
    \centering
\includegraphics[width=0.8\linewidth]{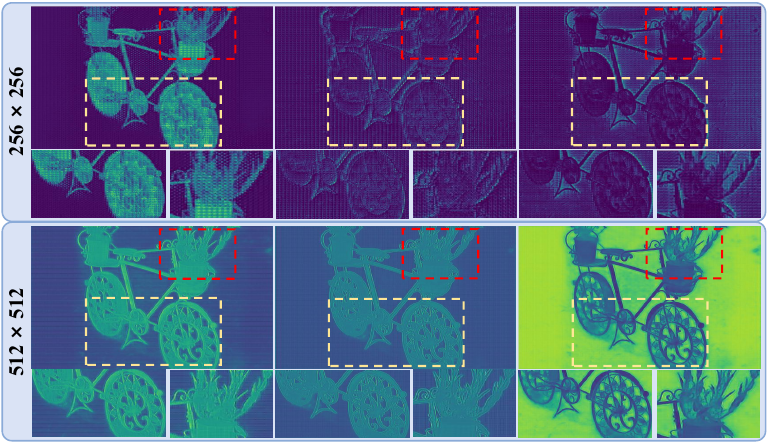}
    \caption{\label{ablation2} \textbf{Comparison of ELE-SAM's feature maps at different resolutions}. It can be seen that the high-resolution feature maps contain clearer object details, thereby facilitating high-quality segmentation.}
\end{figure}

\begin{table*}[htbp]
    \renewcommand{\arraystretch}{1}
    \centering
    \small  
    \caption{\textbf{Comparison of zero-shot segmentation capabilities with SAM and its derivative models on the COCO~\citep{singh2024benchmarkingobjectdetectorscoco} dataset}. FocalNet-DINO~\citep{zhang2022dinodetrimproveddenoising} is used for generating box prompt.}
    \label{zero-shot}
    \setlength{\tabcolsep}{4pt} 
    \begin{tabular}{l|cccccc}
        \toprule
        \textbf{Module} & AP & $\text{AP}_{50}$ & $\text{AP}_{75}$ & $\text{AP}_{L}$ & $\text{AP}_{M}$ & $\text{AP}_{S}$ \\
        \midrule
        SAM~\citep{kirillov2023segment} & 48.5 & 75.5 & 52.7 & 63.9 & 53.1 & 34.1 \\
        SAM-Adapter~\citep{yang2023aimadaptingimagemodels} & 44.8 & 69.5 & 48.1 & 63.9 & 47.8 & 29.0 \\
        SAM-FineTune & 19.5 & 39.1 & 16.2 & 45.2 & 15.8 & 4.7 \\
        HQ-SAM~\citep{sam_hq} & 49.5 & 75.9 & 53.1 & 66.2 & 53.8 & 33.9 \\
        PA-SAM~\citep{xie2024pasampromptadaptersam} & 49.9 & 76.1 & 53.9 & 66.7 & 53.9 & 34.5 \\
        \rowcolor{gray!10}ELE-SAM & \textbf{50.6} & \textbf{76.5} & \textbf{54.2} & \textbf{67.9} & \textbf{54.7} & \textbf{34.8} \\
        \bottomrule
    \end{tabular}
\end{table*}
\subsection{Zero-Shot Comparison with SAM}
To evaluate the generalization capability of our proposed ELE-SAM, we conduct zero-shot segmentation experiment on the COCO\citep{singh2024benchmarkingobjectdetectorscoco} dataset, comparing its performance against SAM and several derivative models. Following HQ-SAM~\citep{sam_hq}, we use the model trained on HQSeg-44K for direct evaluation on COCO in the zero-shot manner. As summarized in Tab.~\ref{zero-shot}, our method demonstrates superior performance across various metrics. Specifically, ELE-SAM achieves 50.6\% Average Precision (AP), outperforming SAM (48.5\%), HQ-SAM (49.5\%), and PA-SAM~\citep{xie2024pasampromptadaptersam} (49.9\%). The results validate that our ELE-SAM also maintain the generalization capability. As illustrated in Fig.~\ref{zero-shot-fig}, ELE-SAM produces fewer artifacts and achieves more accurate segmentation, especially in complex background. It further underscore the robustness of ELE-SAM, making it a strong candidate for real-world applications which require high-quality segmentation without task-specific fine-tuning.

\subsection{Ablation Study}
In this section, we conduct ablation studies on ELE-40K and HQSeg-44K to investigate the effectiveness of High-Fidelity Mask Decoder (HFMD) and Context-Aware Prompt Adapter (CAPA). 
In HFMD, before the fusion process (the same as HQ-SAM~\citep{sam_hq}) between mask features and image fusion features, we plug in another light-weight refinement block (RB). The impacts of the additional refinement block and leveraging high-resolution mask features are discussed. Moreover, we showcase the effectiveness of CAPA over the original Prompt Adapter (PA).

\textbf{Effectiveness of High-Fidelity Mask Decoder.} As shown in Tab.~\ref{tab: effect of hfdm}, using HFMD brings significant performance gains over merely fine-tuning SAM on target datasets. Specifically, compared to SAM-FineTune, incorporating HFMD improves 11.4\% mIoU and 14.5\% mBIoU on ELE-40K, 4.7\% mIoU and 5.4\% mBIoU on HQSeg-44K. Moreover, while comparing HFMD with HFMD (w/o HR) on both datasets, we observe that introducing high-resolution mask features also improves the segmentation results from low-resolution mask features. Besides, comparing HFMD with HFMD (w/o RB), we demonstrate the effectiveness of introducing the light-weight refinement block in HFMD.
More intuitively, we visualize the mask features in the first three channels at 256×256 and 512×512 resolutions in Fig.~\ref{ablation2}. High-resolution mask features capture more fine-grained structure details, benefiting the high-quality segmentation.

\begin{table}[htbp]
    \centering
    \small
    \setlength{\tabcolsep}{3pt}
    \caption{\textbf{Ablation studies on the components of HFMD and CAPA.} `Res' indicates the mask feature resolution used for achieving segmentation outputs. `HR' denotes the usage of high-resolution mask features. `RB' denotes the light-weight refinement block.}
    \label{tab: effect of hfdm}
    \begin{tabular}{l|c|cc|cc}
         \toprule
         & &\multicolumn{2}{c|}{\textbf{ELE-40K}} &\multicolumn{2}{c}{\textbf{HQSeg-44K}} \\
          \multirow{-2}{*}{\textbf{Model}} &\multirow{-2}{*}{\textbf{Res}} &\textbf{mIoU} &\textbf{mBIoU} &\textbf{mIoU} &\textbf{mBIoU} \\
         \midrule
    SAM-FineTune                                  &256 & 65.4    & 56.5   & 88.5    & 80.5     \\
    \midrule
        \multirow{2}{*}{HFMD (w/o HR)} &256 &73.3 &68.4 &90.8 &83.6 \\
         &- &- &- &- &- \\
        \midrule
        \multirow{2}{*}{HFMD (w/o RB) } &256 &68.9 &61.2 &91.7 &82.8 \\
         &512 &70.7 &61.6 &92.2 &85.0 \\
        \midrule
        \multirow{2}{*}{HFMD} &256 &74.8 &69.7 &92.9 &85.7 \\
         &512 &76.8 &71.0 &93.2 &85.9 \\
         \midrule
         \multirow{2}{*}{HFMD + PA} &256 &78.6 &71.3 &93.2 &86.3 \\
         &512 &79.2 &71.4 &93.6 &87.4 \\
         \midrule
         \multirow{2}{*}{HFMD + CAPA} &256 &79.3 &72.9 &92.9 &87.2 \\
         &512 &\textbf{80.0} &\textbf{74.1} &\textbf{94.1} &\textbf{88.3} \\
    \bottomrule
    \end{tabular}
\end{table}

\begin{table}[htbp]
    \centering
    \small
    \setlength{\tabcolsep}{3pt}
    \caption{\textbf{Efficiency influence of different components on ELE-40K.} FPS during inference is reported here with training memory occupation recorded.}
    \label{HFDM efficiency}
    \begin{tabular}{l|cc|cc}
         \toprule
          & \multicolumn{2}{c|}{\textbf{Performance}} & \multicolumn{2}{c}{\textbf{Efficiency}} \\
         \multirow{-2}{*}{\textbf{Model}} & \textbf{mIoU} & \textbf{mBIoU} & \textbf{Memory} & \textbf{FPS} \\
         \midrule
        SAM-FineTune & 65.4 & 56.5 & 16,012MB & 8.31 \\
        HFMD (w/o HR) & 73.3 & 68.4 & 16,014MB & 8.26 \\
        HFMD (w/o RB) & 70.7 & 61.6 & 16,018MB & 7.95 \\
        HFMD & 76.8 & 71.0 & 16,019MB & 7.87 \\
        HFMD + PA & 79.2 & 71.4 & 16,902MB & 7.18 \\
        HFMD + CAPA & \textbf{80.0} & \textbf{74.1} & 16,907MB & 6.85 \\
         \bottomrule
    \end{tabular}
\end{table}

\begin{table}[htbp]
    \centering
    \small
    \setlength{\tabcolsep}{4pt}
    \caption{\textbf{The impact of different backbones on ELE-40K.}}
    \label{Backbone}
    \begin{tabular}{l|cc|cc}
         \toprule
          &\multicolumn{2}{c|}{\textbf{Performance}} &\multicolumn{2}{c}{\textbf{Efficiency}} \\
         \multirow{-2}{*}{\textbf{Backbone}} &\textbf{mIoU} &\textbf{mBIoU} &\textbf{Memory} &\textbf{FPS} \\
         \midrule
        ViT-B &77.6 &68.8 &11,134MB &13.15 \\
        ViT-L &\textbf{80.0} &\textbf{74.1} &16,907MB &6.85 \\
        ViT-H &79.0 &73.2 &24,178MB &5.11 \\
         \bottomrule
    \end{tabular}
\end{table}

\textbf{Effectiveness of Context-Aware Prompt Adapter.} As shown in Tab.~\ref{tab: effect of hfdm}, incorporating the original PA module further enhances the performance, yielding 2.4\% mIoU and 0.4\% mBIoU improvements on ELE-40K. In comparison, replacing PA with CAPA obtains more significant enhancement, which validates the effectiveness of CAPA module for achieving more adaptive and distinctive prompt tokens.

\textbf{Influence of Proposed Modules on Efficiency.} Compared to the fine-tuned SAM, our final ELE-SAM achieves significant performance gains while sacrificing acceptable inference speed. To be concrete, as shown in Tab.~\ref{HFDM efficiency}, ELE-SAM surpasses SAM-FineTune by 16.7\% mIoU and 17.2\% mBIoU on ELE-40K with a cost of 1.46 FPS. We also comprehensively provide the influence of different components on FPS and training memory consumption. As can be seen, these components are effective on performance and efficient on computation.

\textbf{Impact of Different Backbones.} When changing the size of ViT backbone, we find that ViT-L obtains the best performance on ELE-40K. The detailed metrics are listed in Tab.~\ref{Backbone}. While ViT-L outperforming ViT-B by a clear margin, the larger frozen ViT-H does not further promote the performance. It could be attributed to the over-smoothing issue~\citep{ru2023token} for deeper ViT.

\begin{figure}[!t]
    \centering
    \includegraphics[width=1\linewidth]{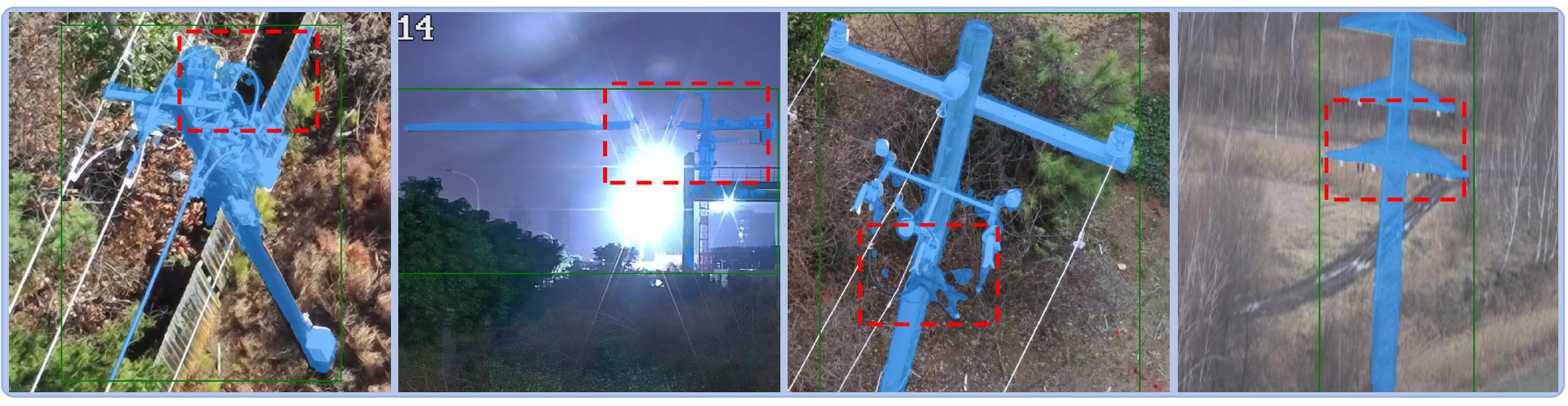}
    \caption{\label{fig11}\textbf{Analysis on challenging cases.} Example failure cases under highly challenging conditions (complex occlusions, low illumination, and adverse weather), which present opportunities for future robustness enhancement.}
\end{figure}

\section{Limitation and Discussion}
\label{sec: limitation}
Although ELE-SAM achieves superior performance on the PTCHS task and general segmentation benchmarks, there is still potential for further improvement. As illustrated in Fig.~\ref{fig11}, the segmentation quality may degrade under highly challenging conditions such as occlusion, low illumination, and adverse weather. 
While ELE-40K encompasses these scenarios, further architectural refinements can be developed to address these challenging cases.

In addition, as demonstrated in Tab. \ref{HFDM efficiency} and Tab. \ref{Backbone}, ELE-SAM currently cannot achieve real-time segmentation based on large image encoders, such as ViT-L and ViT-H. Future work could leverage lightweight vision backbones to speed up inference and explore better multi-scale feature aggregation schemes to further promote the segmentation quality.
\section{Conclusion}
\label{sec: conclusion}
In this paper, we present ELE-SAM, an effective solution for the Power Transmission Corridor Hazard Segmentation (PTCHS) task. Two key modules named Context-Aware Prompt Adapter (CAPA) and High-Fidelity Mask Decoder (HFMD) are designed to address the challenges posed by complex backgrounds and heterogeneous object structures. 
CAPA mines more discriminative prompt tokens for better distinguishing target objects from background, while HFMD segmenting them and preserving high-fidelity structure details by scaling up the mask features to a higher resolution. 
To further promote the research field, we contribute a large-scale benchmark named ELE-40K, including 44,094 image-mask pairs covering 4 electric power transmission equipments and 11 hazard categories.
According to the experiments, ELE-SAM achieves state-of-the-art performance on ELE-40K. Moreover, we also demonstrate the effectiveness and generalization of our method on high-quality segmentation for general objects.

\section*{Acknowledgements}
This work was financially supported by the State Grid Corporation Headquarters Science and Technology Project: Research on equipment operation and inspection disposal reasoning technology based on knowledge-enhanced generative model and intelligent agent and demonstration application (5700-202458333A-2-1-ZX). The numerical calculations in this paper have been done on the supercomputing system in the Supercomputing Center of Wuhan University.

\section*{Data availability.} 
All datasets used in this paper are publicly available. ELE-40K can be accessed at https://github.com/Hhaizee/ELE-SAM. HQSeg-44K~\citep{sam_hq} can be accessed at https://github.com/SysCV/SAM-HQ. COCO~\citep{singh2024benchmarkingobjectdetectorscoco} can be accessed at https://github.com/cocodataset/cocoapi.

\bibliographystyle{elsarticle-num}
\bibliography{bib}
\end{document}